\documentclass{article}

\usepackage{arxiv}

\usepackage[utf8]{inputenc} 
\usepackage[T1]{fontenc}    
\usepackage{hyperref}       
\usepackage{url}            
\usepackage{booktabs}       
\usepackage{amsfonts}       
\usepackage{nicefrac}       
\usepackage{microtype}      
\usepackage{lipsum}

\usepackage{fancyhdr,graphicx,amsmath,amssymb,mathtools}

\title{Odds-Ratio Thompson Sampling to Control for Time-Varying Effect}

\author{
 Sulgi Kim \\
  Clova AI, NAVER Corp.\\
  \texttt{sulgi.kim@navercorp.com} \\
   \And
 Kyungmin Kim \\
  Clova AI, NAVER Corp.\\
  \texttt{kyungmin.kim,ml@navercorp.com} \\
}  
  
\begin{document}
\maketitle

\begin{abstract}
Multi-armed bandit methods have been used for dynamic experiments particularly in online services. Among the methods, thompson sampling is widely used because it is simple but shows desirable performance \cite{agrawal2012analysis, chapelle2011empirical}. Many thompson sampling methods for binary rewards use logistic model that is written in a specific parameterization. In this study, we reparameterize logistic model with odds ratio parameters. This shows that thompson sampling can be used with subset of parameters. Based on this finding, we propose a novel method, ``Odds-ratio thompson sampling'', which is expected to work robust to time-varying effect. 
Use of the proposed method in continuous experiment is described with discussing a desirable property of the method. 
In simulation studies, the novel method works robust to temporal background effect, while the loss of performance was only marginal in case with no such effect. Finally, using dataset from real service, we showed that the novel method would gain greater rewards in practical environment.
\end{abstract}

\section{Introduction}
Experimentation platform is now an essential part in online services for providing the best service by evaluating and comparing multiple candidates in real services \cite{scott2015multi}. 
A/B test and multi-armed bandit are main methodologies that direct a design and a decision in an experiment.
Multi-armed bandit methods provide simple but flexible experiment framework compared to A/B test \cite{scott2015multi, burtini2015survey}.
For example, thompson sampling \cite{thompson1933likelihood}, a popular method of multi-armed bandit, outputs a result of experiment in terms of allocation of next experiment, which allows a sequential experiment \cite{russo2018tutorial}. 

In real service, it is common to have a type of non-stationary environment, that is time-varying effect \cite{scott2015multi, canamares2019multi, raj2017taming, zeng2016online, burtini2015improving, gupta2011thompson}. However, multi-armed bandit including thompson sampling is sensitive to this irregular condition in nature compared to A/B test, where sample sizes for all variants do not change during an experiment.
In a continuous experiment where arms are added and dropped in the middle of the experiment, the time-varying effect would cause more severe sub-optimal result of the multi-armed bandit policy. It is because each arm is tested in different period.

In this study, we propose a natural way of dealing with time-varying effect with reparameterizing base model for thompson sampling. First, we will describe different parameterizations of logistic model and propose a novel thompson sampling (Odds-ratio thompson sampling) policy with a specific parameterization. Then its use in continuous experiment will be described. Finally we will evaluate its empirical regret together with other methods in simulated data.

\section{Problem Setting}
Many multi-armed bandit applications adopt \textit{Batch Update}, where arms are played multiple times, then, policy and related parameters are updated with aggregated rewards \cite{chapelle2011empirical, russo2018tutorial}. Batch update, which is sometimes called as delay update, is practical set up, 
because it requires much less computational resources than online or realtime update. 
There are many chances that the temporal effect changes concurrent with batch update, making reward probabilities change through rounds.
Our study discusses thompson sampling that are applied with the batch update, while there is common effect in each round.

Mean reward of $i$-th arm is represented by $E(R_i)=p_i$, where $R_i$ is a random variable for binary reward. Later we will discuss these mean rewards, or performances can drift through rounds.
Reward observations are aggregated at each round and given as $r_i = (n_i, c_i)$, where $n_i$ is the number of trials of $i$-th arm and $c_i$ is its number of successes. For simplicity of exposition, we suppress a subscript $t$ if it is not necessary. For a task finding the arm with the maximum click through rate (CTR), page view counts (or impression) and click counts can be used for $n_i$ and $c_i$, respectively. 

In batch update, we allocate samples (or traffics) to each arm according to their thompson probabilities, instead of choosing actions with their computed probabilities. In the context of batch update, we call thompson probability as \textit{proportion}.

\section{Full Rank Thompson Sampling}
In this section, we will parameterize logistic model, and introduce the first version of thompson sampling policy based on the logistic model, which we denote as full rank thompson sampling (\textbf{Full-TS}).
Logistic model for multi-armed bandit for $K$ number of arms is given:
\begin{equation}
\log \left({p_i}\over{1-p_i}\right)= \sum \beta_i x_i +\beta_K \label{eq:1}
\end{equation}
, where $x_i=\left\{\begin{matrix*}[l] 1 &\text{, if arm $i$ is pulled}\\ 0 &\text{, otherwise} \end{matrix*}\right.$ 

We can see each parameter except $\beta_K$ is written as:
\begin{displaymath}
  \beta_i = \log\left( {p_i\over{1-p_i}} / {p_K\over{1-p_K}} \right) \text{for } i=1, \cdots, K-1
\end{displaymath}
, which represents (log of) \textit{Odds Ratio}, $i$-th arm's reward probability with respect to reference $K$-th arm in logit scale. 
Note $\beta_i > 0 $ if and only if $p_i > p_K$. 
We denote parameter as follows:
\begin{displaymath}
  \Tilde{\beta} = \begin{pmatrix}\Tilde{\beta}_{OR}\\
\beta_K\end{pmatrix} ,
\Tilde{\beta}_{OR} = \begin{pmatrix}\
    \beta_1 \\
    \vdots \\
    \beta_{K-1}
\end{pmatrix} 
\end{displaymath}

Our logistic model, \eqref{eq:1} is written differently from typical setting, where $\beta_K$ in \eqref{eq:1} is replaced by $\beta_K^\prime x_K$. Let us denote this parameterization as $\Tilde{\beta}_{ind}$. 
In fact, the two parameterizations have a relationship of linear transformation,  $\Tilde{\beta}_{ind} = C_{ind}\Tilde{\beta}$, where $C_{ind}$ is $K\times K$ matrix with entries,
\begin{equation}
\label{eq:C}
    {C_{ind}}_{i,j} = \left\{\begin{matrix*}[l]
    1 &\text{ , if $i=j $ or $ j = K$}\\
    0 &\text{ , otherwise}
    \end{matrix*}
    \right. 
\end{equation}

This transformation does not change the model's representation, thus the two parameterizations are equivalent to each other.

Once our thompson sampling begins with initial proportion for each arm, it alternates two steps: posterior update and allocation proportion, each step of which is described in detail in separate sections.

\subsection{Prior and Initial Proportions}
We use non-informative, uniform prior, which is an imaginary distribution that has constant density on ${\rm I\!R}$. 
We found use of this improper prior has many advantages, described below.

We manually give initial proportions, $p_{TS, i}$'s as $1/K$. 
We found setting $\Sigma_0^{-1}$ as zero-matrix from $\Tilde{\beta}\sim N(\mu_0, \Sigma_0)$ makes the following computation simpler.

\subsection{Posterior Update}
Thompson sampling begins with prior, then as rewards are observed, posterior is updated using bayes rule:
\begin{displaymath}
P(\Tilde{\beta}|r) = P(r|\Tilde{\beta})P(\Tilde{\beta})
\end{displaymath}
, where $P(\Tilde{\beta})$ is posterior at previous round, or prior if $t=1$, $P(r|\Tilde{\beta})$ is likelihood and $P(\Tilde{\beta}|r)$ is posterior that is updated in current round. For given round, $t$, we will call $P(\Tilde{\beta})$ as prior at the round. 

We follow a general Bayesian logistic regression described in \cite{bishop2006pattern}. Each time we will do Laplacian approximation to keep posterior as Gaussian distribution:
\begin{displaymath}
  \Tilde{\beta}|r \sim N\left(\mu_t, \Sigma_t\right)
\end{displaymath}
, where $\mu_t$ and $\Sigma_t$ is mean and covariance parameter of Gaussian distribution, respectively. Note that our parameters, $\beta_i$s are not independent to each other, so posterior is multivariate distribution that is not factorized into univariate distributions.

Negative log of posterior is 
\begin{equation}
\label{eq:posterior}
  l_t(\mu) = - {1\over{2}} \left({l(r|\mu) + l(\mu, \Sigma_{t-1})}\right)
\end{equation}
, where $l(r|\mu)$ is log of binomial likelihood and $ l(\mu, \Sigma_{t-1})$ is log of prior, that is log probability density function of Gaussian distribution.

The two parameters are updated sequentially,
\begin{equation}
\label{eq:update}
\begin{matrix}
\mu_t = \text{argmin}_{\mu}l_t(\mu) \\
\Sigma_t^{-1} = \Sigma_{t-1}^{-1} + \Lambda
\end{matrix}
\end{equation}
, where $\Lambda$ is the second derivative of \eqref{eq:posterior} at $\mu = \mu_t$. 
Using $\hat{p}_i$ which is derived from $\mu_t$, $\Lambda$ is represented with its $(i,j)$-th entry, 
\begin{displaymath}
\Lambda_{ij} = 
\left\{\begin{matrix*}[l]
n_i\hat{p}_i(1-\hat{p}_i) & \text{, for $i=j\ne K$, or $i = K \ne j$, or $j =K \ne i$  }\\ 
\sum_{i=1}^{K} {n_i\hat{p}_i(1-\hat{p}_i)} & \text{, for $i=j=K$  }\\ 
0 &\text{, otherwise} 
\end{matrix*}\right.
\end{displaymath}

\subsection{Getting Allocation Proportion} \label{proportion}
Order among mean reward of arms is determined by order among $\beta_i$'s $(i=1,\cdots, K-1)$ and $0$. That is, we set
\begin{displaymath}
  \text{Set } \beta^{\prime}_i =\left\{
  \begin{matrix*}[l]
  \beta_i & \text{, for } i = 1, \cdots, K-1\\
  0 &\text{, for } i = K
  \end{matrix*}\right.
\end{displaymath}

Then, Thompson Sampling states probability with which each arm is pulled (or selected), $p_{TS, i}$ is given,
\begin{equation}
\label{eq:allocation}
    p_{TS, i} =\int I\left(i = \text{argmax}_j \beta_j^{\prime} \right) P(\Tilde{\beta}|r)d\Tilde{\beta}  
\end{equation}
, where $I$ is an indicator function, $p_j\left(\Tilde{\beta}\right)$ is a function that maps $\Tilde{\beta}$ to $p_j$, which is derived from \eqref{eq:1}.
This can be obtained at round $t$ by generating $N$ number of multivariate samples $\Tilde{\beta}^s$ from $ N\left(\mu_t, \Sigma_t\right)$ and compute $i$-th arm's proportion,
$p_{TS, i} = {1 \over N} \sum I(i = \text{argmax}_j (p_j(\Tilde{\beta}^s)))$ then, randomly allocate $i$-th arm to $p_{TS,i}$ proportion of next round.

\subsection{Linear Transformation and Invariance}
In our logistic model, it may be less intuitive whether the result is affected by an indexing of arms, for example, changing a reference arm. 

Any indexing of same arm set can be transformed to the independent parameterization, $\Tilde{\beta}_{ind}$, and its posterior is accordingly transformed to $\Tilde{\beta}_{ind}\sim N(C_{ind}\mu, C_{ind}\Sigma C_{ind}^T)$. Once transformed to the parameterization, univariate parameter in $\Tilde{\beta}_{ind}$ is independent to each other, where it is obvious different indexings give the identical posterior. 

We call this property as "reference-invariance" property. In other domain, it has been understood that different encoding of categorical variable does not matter. Here, we see that this is true for multi-armed bandit using the logistic model with uniform prior.

\section{Odds-Ratio Logistic Thompson Sampling}
In previous section, the logistic model \eqref{eq:1} has the interesting property: ''Order among reward probability $p_i$'s is determined only by comparing among $\Tilde{\beta}_{OR}$ and 0, rather than $\Tilde{\beta}$.'' 
In other words, parameter required for step \eqref{eq:allocation} is $\beta_{OR}$, and $\beta_K$ does not contribute any information. 

This property motivated us to devise a bandit model where each round shares only OR parameters $\Tilde{\beta}_{OR}$, rather than full $K$ parameters, $\Tilde{\beta}$. We name this model as Odds-Ratio Thompson Sampling (\textbf{OR-TS}). Freeing intercept parameter at each round corresponds to allow each arm's reward probability drift of the same interval, in logit scale at each round. 

We describe Odds-Ratio Logistic Bandit. Posterior update at round $t-1$ would produce posterior for $\beta_{t-1}$ to be used as prior at round $t$.
Then updating only odds-ratio parameters can be symbolically described as changing the equality from $\beta_{K,t-1}=\beta_{K,t}$ to $\beta_{K,t-1} \ne \beta_{K,t}$. 

We can obtain prior for $\beta_{OR,t}$ as $(k-1)$-variate Gaussian distribution, $N(\mu_{-K}, \Sigma_{-K})$ by marginalizing full rank prior with respect to $\beta_{K,t-1}$, where $\mu_{-K}$ and $\Sigma_{-K}$ are
\begin{equation}
    \label{eq:marginalize}
    \begin{matrix*}[l]
    &\mu_{-K} &=& \mu_{(:K-1)} \\
    &\Sigma_{-K} &=& \Sigma_{(:K-1, :K-1)}
    \end{matrix*}
\end{equation}

We use a uniform prior for $\beta_{K,t}$ as we have done for parameters at $t=1$. Then, bayes rule for posterior update is written:
\begin{equation} \label{eq:orupdate}
    P(\Tilde{\beta}|r) = P(r|\Tilde{\beta})P(\Tilde{\beta}_{OR})
\end{equation}
Once we replace $l(\mu, \Sigma_{t-1})$ of \eqref{eq:posterior} by $l(\mu_{-K}, \Sigma_{-K, t-1})$, we follow the same downstream process as in Full-TS.

It is worth to noting that this marginalization does not depend on a reference arm. In other words, OR update also has a \textit{reference-invariance} property. 
This can be seen by the fact that any indexed marginalized prior is transform to independent parameterization, that is identical (degenerate) Gaussian distribution. 

Note that OR-TS and Full-TS is defined for a round, thus, one can switch using between OR-TS or Full-TS at each round through an experiment.
That is, at one round, assuming $\beta_{K, t-1} \ne \beta_{K,t}$ and at other round, $\beta_{K, t-1} = \beta_{K,t}$. 

\section{Application in Continuous Experiment}
Multi-Armed Bandit has been discussed with strict setting, where an experiment period or an arms set is assumed to be fixed during an experiment. 
However, multi-armed bandit can be used in more flexible way, such as without fixing the experiment period beforehand \cite{canamares2019multi}.
We define continuous experiment, the one where a set of arms to be allocated (denoted by $A_t$) changes over rounds and its period is not fixed.
Simple scenarios include a case when new arms are added in the middle of an experiment. 

Odds Ratio Thompson Sampling is more desirable in the continuous experiment, because periods for which each arm has been observed, differ, thus non-stationary environment affects full rank bandit worse. 

We denote a set of arms observed until round $t$ as $C_t$. Generally it can be represented as cumulative set: $C_t = A_1\cup \cdots \cup A_t$. 
For simplicity, we assume all arms in $A_t$ are pulled (i.e. observed) at round $t$. 
Note that information from previous observations is delivered only when 
\begin{equation}
\label{eq:condition}
  n\left(A_t \cap C_{t-1}\right)\ge2
\end{equation}
, where $n(\cdot)$ is the number of items in a group. 
Therefore, to continue OR bandit, one should design $A_t$ satisfying the condition, \eqref{eq:condition}. Nevertheless, when $n\left(A_t \cap C_{t-1} \right)=1$, we can just begin new multi-armed bandit by initializing $\mu$ and $\Sigma$, or we can use full rank prior.

\subsection{Getting Allocation Proportions in Continuous Experiment}
Parameter, its prior and posterior can be freely transformed for any indexing of arms due to the reference-invariance property. This transformation is required when a previous reference arm is not included in $A_t$, for example, because it is removed due to its low performance. In this case, we need to change a reference arm to one from $A_t \cap C_{t-1}$ and reparameterize it accordingly. 
Let $f$ be the mapping function, that is, $f(i)$ is a new index for previous $i$-th arm. 
The transform matrix, $C$ can be represented as a chain of transformations: $C = {C_{ind}}^{-1} C_f C_{ind}$, where $C_f$ is corresponding matrix for $f$, with entries:
\begin{displaymath}
    {C_f}_{i,j} = \left\{\begin{matrix*}[l]
    1 &\text{, if $j = f^{-1}(i)$}\\
    0  &\text{, otherwise}
    \end{matrix*}
    \right. 
\end{displaymath}

Getting allocation proportions for $A_t$ in a continuous experiment is straightforward using the rules previously applied. Assume that $A_t$ comprises arms already observed, thus in $C_{t-1}$ and the new ones which is not in $C_{t-1}$. Denote the two groups as $O_t$ and $U_t$, respectively. First, we manually set proportions for arms in $U_t$ as $1/n(A_t)$. We transform posterior from indexing in posterior to new indexing for arms in $O_t$. Then, we can obtain $p_{TS,i}$s for $i$-th arm in $O_t$ on transformed posterior as described in section \ref{proportion}. Then, allocation proportions for arms in $O_t$ is given ${{n(O_t)}\over{n(A_t)}} p_{TS,i}$. Note this step is identical in Full-TS.
\subsection{Posterior Update in Continuous Experiment}
Posterior for $C_{t-1}$ at round $t-1$ is updated as we observe $r$ for $A_t$. Again, if a reference arm in a given parameterization of posterior at previous round, $t-1$ is not in $A_t$, change the reference arm to one in $A_t \cup C_{t-1}$, and transform parameterization accordingly. Note that in previous allocation step, we transform to arms in $A_t$, but for posterior update we transform posterior to all arms in $C_t$. This previous posterior is marginalized in terms of reference arm, in equation \ref{eq:orupdate}. Since all parameters are correlated, parameters for arms in $U_t$ are also updated. As far as $C_t$ grows with satisfying \eqref{eq:condition} at round $t$, then posterior for all arms in $C_t$ are successfully accumulated. 

Information about relative performance among arms in $C_t$ is efficiently summarized in posterior. Note that this includes arm pair which has not been directly observed at the same round. For example, assume $A_1= \{A,B,C\}, A_2= \{B,C,D\}$. Even though two arms, $A$ and $D$ are not directly compared at the same round, indirect information from their odds ratios with respect to a common reference, that is, B or C, is summarized in posterior.

\section{Experiment}
\subsection{Simulation Study}
We compared three different thompson sampling policies for binary reward: Full-TS, OR-TS and Beta-Bernoulli Thompson Sampling (\textbf{Beta-TS}). We simulated various environments and investigated behaviours of the three policies. 

We set $K=10$ arms, one of which has greater mean reward than the other nine arms. In specific, $\beta_i$ for nine sub-optimal arms are set to have $p_i = 0.30$, and $0.31$ for an optimal arm. Then at each round, all arms have common background effect, $\delta_t$, which is generated from Gaussian distribution:
\begin{displaymath}
    \begin{matrix}
    \beta_{i,t} = \beta_i + \delta_t \\
    \delta_t \sim N(0, \sigma^2)
    \end{matrix}
\end{displaymath}
Level of time-varying effect is controlled by variance of Gaussian distribution, $\sigma$. 
We represent the level of time-varying effect by a scaled metric, $d$ which represents relative scale to mean reward difference between optimal and sub-optimal arms.
We repeated simulation 100 times with 50 rounds for each simulation. Each round consists of 10,000 trials.

\begin{figure}
    \centering
    \includegraphics[scale=0.55]{{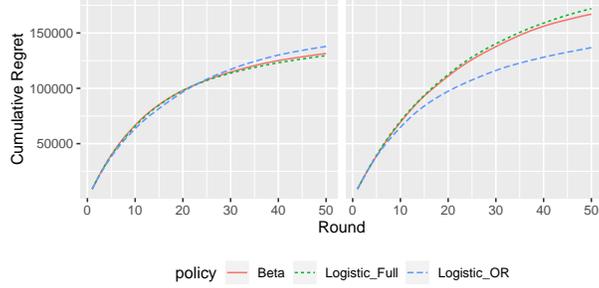}}
    \caption{Cumulative Regrets for different level of time-varying effect ($d$): Left panel for no effect ($d=0$) and right panel for with time-varying effect ($d=20$). Total sample size per round is set to $N=10,000$. }
    \label{fig:sidebyside}
\end{figure}

Mean of cumulative regret is shown in \textbf{Figure \ref{fig:sidebyside}}. 
For both panels, we see that Beta-TS and Full-TS has very similar regret as expected. 
In environment of no time-varying effect (left panel), OR-TS have a slightly greater regret than the other two policies. 
However, when there is time-varying effect (right panel), OR-TS is robust to the effect, while Beta-TS and Full-TS are affected severely.
\subsection{Experiment Based on Real Digital Advertisement Data}
We investigated the performance of the three policies with real digital advertisement data of online messenger users. 
Even number of users have been exposed to four different listings which is sub module in a mobile page for 18 days. 
Setting an arm as showing each listing, we consider a multi-armed bandit problem of finding the best listing that has the maximum CTRs among the four listings. 
On the $11$-th day of the experiment, the listings are shifted up into very top screen. Due to this change, performance values of the arms have been changed: page views have increased more than two times, while CTRs have decreased because of increase of views. 
Therefore, the data is expected to have two types of varying effects: one minor effect daily and the other major effect between the two periods.

\begin{figure}
    \centering
    \includegraphics[scale=0.52]{{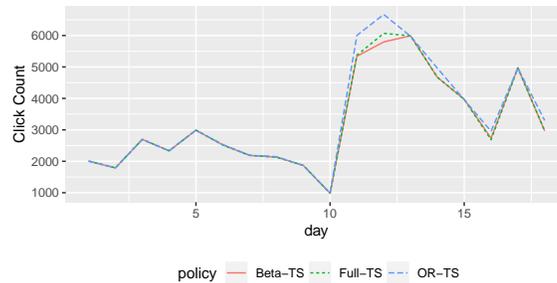}}
    \caption{Expected Click Counts for Advertisement Data}
    \label{fig:line}
\end{figure}

We set ground truth CTRs as estimated from the real data and simulated multi-armed bandit.
To investigate behaviors of the policies more efficiently, we assume total sample size is about $20,000$ to $200,000$ page views per round, which amounts to $20\%$ of real data.
Then, we record expected rewards, i.e. click counts for each round from the three policies.
\textbf{Figure \ref{fig:line}} shows click counts for Beta-TS, Full-TS, and OR-TS. 
From the figure, indirectly we see time-varying effect through the period.
OR-TS has gained more clicks compared to the other policies, especially during the two days after a major event occurred.
We may infer that the temporal effects in day 1-10 (especially low performance in $10$-th day) resulted in biased posterior for Beta-TS and Full-TS, while OR-TS worked robust to the effects. During the period, it is estimated that OR-TS gains $3.9\%$ more clicks than Beta-TS and Full-TS.


\section{Conclusion and Discussion}
In this study, we investigated an alternative expression of logistic model for thompson sampling. Based on the expression we showed that Odds-Ratio only update makes multi-armed bandit policy robust in many practical environment. Therefore, we believe that one may consider Odds Ratio thompson sampling as standard policy for binary reward data in batch update settings.

This study focused on the case when there is a common background effect, which does not change an optimal arm. However, real data can be confounded with the fact that an optimal arm changes over time. In this case, use of plain OR-TS would be also affected. In this case, we can use discount TS \cite{raj2017taming} or aggressiveness parameter \cite{scott2015multi}. For example, we can use multiply $(1-\lambda)$ in second term of \eqref{eq:1}, where $\lambda$ is a decay parameter ranging $\left[0, 1\right]$.

We have discussed a base logistic model. Its extension to other generalized linear model, or contextual bandit is straightforward. Our implementation of OR-TS and Full-TS is available at  \textit{http://github.com/sulgik/orts}.

\bibliographystyle{unsrt}
\bibliography{sulgik}

\end{document}